\documentclass[conference]{IEEEtran}
\IEEEoverridecommandlockouts
\usepackage{cite}
\usepackage{amsmath,amssymb,amsfonts}
\usepackage{algorithmic}
\usepackage{graphicx}
\usepackage{textcomp}
\usepackage{xcolor}
\usepackage{multirow}
\usepackage{hyperref}
\usepackage{float}
\usepackage{stfloats}
\def\BibTeX{{\rm B\kern-.05em{\sc i\kern-.025em b}\kern-.08em
    T\kern-.1667em\lower.7ex\hbox{E}\kern-.125emX}}
\begin{document}

\title{An Efficient Machine Learning Approach for Degradation Forecasting in AEM Water Electrolysis}

\author{
\IEEEauthorblockN{Marco Veneriano}
\IEEEauthorblockA{\textit{MaLGa Center} \\
\textit{Università degli Studi di Genova}\\
Genoa, Italy \\
marco.veneriano@edu.unige.it}
\and
\IEEEauthorblockN{Ani Gjergji}
\IEEEauthorblockA{\textit{MaLGa Center} \\
\textit{Università degli Studi di Genova}\\
Genoa, Italy \\
ani.gjergji@edu.unige.it}
\and
\IEEEauthorblockN{Sebastiano Bellani}
\IEEEauthorblockA{\textit{Antares Electrolysis S.r.l.}\\
Genoa, Italy \\
sebastiano.bellani@antares-electrolysis.com}
\and
\IEEEauthorblockN{Andrea Riva}
\IEEEauthorblockA{\textit{Antares Electrolysis S.r.l.}\\
Genoa, Italy \\
andrea.riva@antares-electrolysis.com}
\and
\IEEEauthorblockN{Vito Paolo Pastore}
\IEEEauthorblockA{\textit{MaLGa Center} \\
\textit{Università degli Studi di Genova}\\
Genoa, Italy \\
vito.paolo.pastore@unige.it}
\and
\IEEEauthorblockN{Matteo Santacesaria}
\IEEEauthorblockA{\textit{MaLGa Center} \\
\textit{Università degli Studi di Genova}\\
Genoa, Italy\\
matteo.santacesaria@unige.it }
\thanks{This is the accepted version of a paper to appear in the Proceedings of the 2026 2nd International Conference on AI and Emerging Technology for Sustainable Future (ICAISF), Catania, Italy. \copyright~2026 IEEE. Personal use of this material is permitted. Permission from IEEE must be obtained for all other uses, in any current or future media, including reprinting/republishing this material for advertising or promotional purposes, creating new collective works, for resale or redistribution to servers or lists, or reuse of any copyrighted component of this work in other works.}
}

\maketitle

\begin{abstract}
This study provides a data-driven analysis of a novel dataset of single-cell Anion Exchange Membrane water electrolyzers (AEMWE), operated under constant current load across multiple heterogeneous experimental campaigns.
We train and evaluate a range of machine learning models with different complexity, including linear baselines, LSTMs and CNNs, to perform medium-term forecasting of the cell voltage degradation curve. The models are assessed within a rigorous training and evaluation framework specifically designed for heterogeneous industrial data.

\end{abstract}


\section{Introduction}
Green hydrogen production via water electrolysis is widely regarded as a key technology for reducing emissions in both industrial and civilian sectors.
Anion Exchange Membrane Water Electrolysis (AEMWE) has recently emerged as a promising alternative to conventional Proton Exchange Membrane (PEM) and alkaline systems. AEMWE offers the potential for lower costs by reducing the reliance on fluorinated polymers and precious metals, while also mitigating their associated environmental impact.
However, a major limitation for large-scale deployment is the limited long-term stability of current membranes, often limiting system lifetime. For this reason, devising systems capable of robust predictive health and performance monitoring is crucial to optimize maintenance and avoid critical failures.
In this work, we present the design and evaluation of a data-driven framework, applied to an in-house AEMWE single-cell dataset, with the goal of developing predictive models for degradation dynamics on a budget, and providing insights applicable to industrial settings.

In recent years, a growing body of work has explored data-driven degradation modeling in electrochemical energy systems, primarily focusing on PEM fuel cells and PEM water electrolyzers. Most of these works focus on short-term prediction or remaining useful life estimation (RUL), often using deep architectures such as CNN--LSTM or attention-based models \cite{b1, b2, b3, b4}.
AEMWE-related machine learning studies have mainly addressed performance prediction and operating-point optimization \cite{b6, b11}, rather than explicit forecasting of degradation trajectories over time, with only a limited number of works addressing it\cite{b10}.
Complementary to data-driven approaches, numerous physics-based models have been developed to describe electrochemical performance and degradation mechanisms in water electrolyzers, typically grounded in electrochemical kinetics, transport phenomena, and material aging laws \cite{b12}. For PEM water electrolyzers, several works review and develop degradation modeling strategies that combine mechanistic descriptions of membrane thinning, catalyst dissolution, and gas crossover with simplified polarization equations to predict efficiency loss and lifetime under realistic operating conditions \cite{b13}.
Building on these concepts, recent studies propose layer-unspecific physical models that extract quasi-steady polarization curves from operating data and track the temporal evolution of effective resistance and exchange current density, yielding millivolt-scale voltage prediction errors and enabling lifetime forecasts with uncertainty quantification \cite{b8}. 
Hybrid approaches have recently been adopted, like Physics-Informed Neural Networks (PINNs). These provide an alternative that improves consistency and interpretability through electrochemical and transport priors, such as temperature prediction for PEM cells \cite{b9}. For AEMs hybrid models are adopted in \cite{b14}, to focus on the evolution of hydroxide conductivity under prolonged alkaline exposure as key indicator of chemical and structural breakdown.
While these physics-based models offer interpretability and extrapolation capabilities, they require extensive characterization of material properties, careful calibration, and often assume quasi-stationary or slowly varying operating conditions \cite{b7}.




Considering a heterogeneous dataset of AEMWE single-cell aging campaigns provided by Antares Electrolysis, this study aims to model AEMWE degradation by forecasting the future evolution of the voltage curve over medium-term horizons. The dataset comprises 12 distinct experimental runs, with different setup configurations (e.g. membranes, catalyst layers, etc.), each lasting between 40 and 100 hours under different configurations. Importantly, the time series considered correspond to the initial phase of operation, characterized by transient dynamics and significantly higher degradation rates compared to steady mid-life conditions. As a result, this makes the forecasting task particularly challenging and distinct from typical degradation studies, which often focus on quasi-stationary regimes \cite{b1, b2, b3}. 

Accordingly, the adopted modeling choices are driven by the need for architectures that can generalize to unseen experimental conditions, remain robust to noise, and perform effectively in a data-scarce regime. Specifically, we formulate a multi-horizon forecasting problem in which a 24-hour observation window of voltage measurements is used to predict the future voltage at horizons of 3, 6, 12, 18, and 24 hours.

To this end, the main contributions of this work are summarized as follows:
\begin{itemize}
    \item We introduce and analyze a unique dataset of AEMWE degradation trajectories, capturing transient early-life dynamics across multiple heterogeneous experimental runs.
    \item We propose a rigorous framework for training and evaluation in data-scarce settings, featuring a Leave-One-Group-Out cross-validation protocol specifically tailored for industrial deployment.
    \item We systematically compare machine learning models of varying complexity, ranging from linear and non-linear baselines to shallow regularized networks and deep temporal architectures (LSTMs, 1D-CNNs), to identify the optimal trade-off between expressive power and stability.
\end{itemize}

\section{Experimental Dataset and Problem Definition}
\label{sec:dataset}

\subsection{Experimental Dataset and Notation}

The experimental analysis is based on a dataset of electrochemical time series provided by Antares Electrolysis, obtained from constant-current aging campaigns on a 5~cm$^2$ Anion Exchange Membrane (AEM) single cell. These tests were performed under heterogeneous operating protocols using a custom test station specifically engineered to characterize cell degradation behavior. The complete dataset comprises 12 distinct experimental runs, each lasting between 40 and 100 hours, with a primary focus on capturing cell voltage evolution over time.

Formally, the dataset is defined as a collection of $N$ experiments:
\begin{equation}
\mathcal{D} = \{E_1, \dots, E_N\},
\end{equation}
where each individual experiment $E_i$ is represented by the tuple:
\begin{equation}
E_i = \left( \{(t_k^{(i)}, v_k^{(i)})\}_{k=1}^{K_i}, \ \mathbf{m}^{(i)} \right),
\end{equation}
where $K_i$ is the number of samples of experiment $i$.
Here, $t_k^{(i)} \in [0, \mathcal{T}_i]$ denotes the discrete measurement timestamp, $v_k^{(i)} \in \mathbb{R}$ represents the observed cell voltage at that timestamp, and $\mathbf{m}^{(i)} \in \mathbb{R}^{d_m}$ is a vector of static metadata detailing the specific initial experimental conditions of the run (e.g. membrane properties, catalyst layers, etc.).

\begin{figure}
    \centering
    \includegraphics[width=1.0\linewidth]{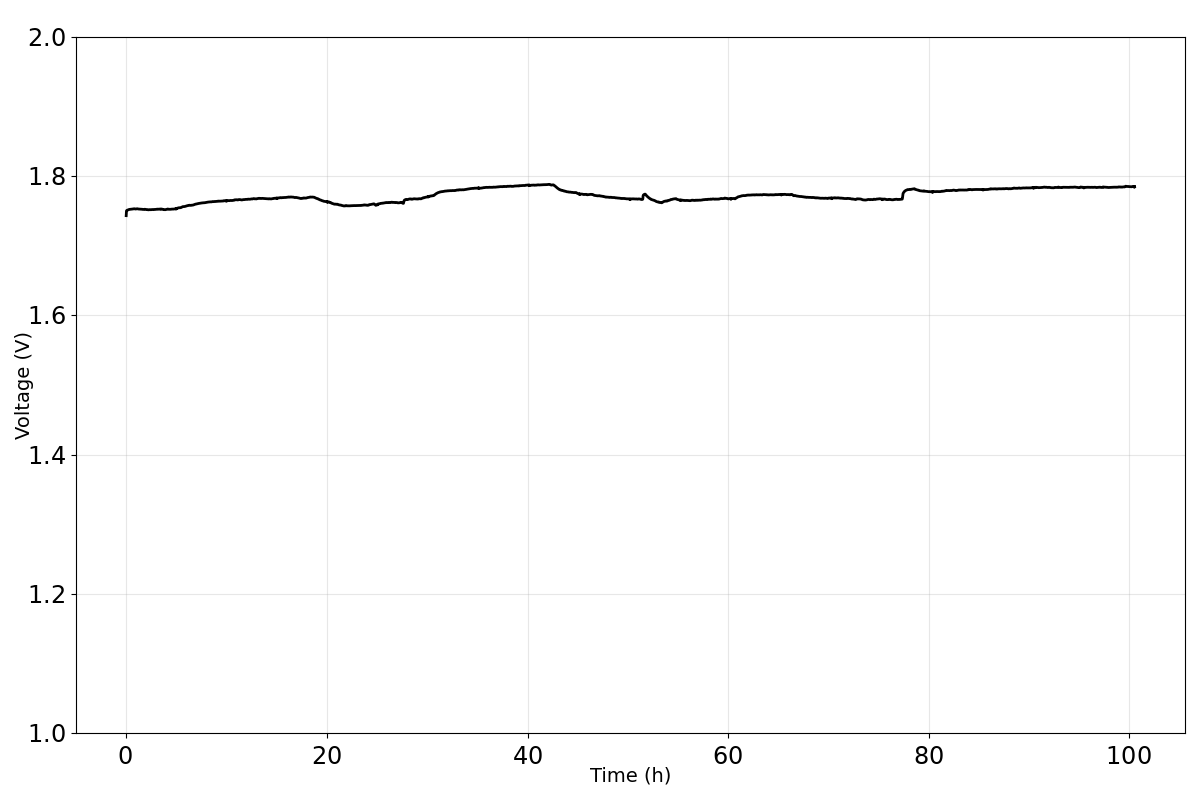}
    \caption{Representative raw voltage profile from a single experiment.}
    \label{figure_experiment}
\end{figure}

A representative raw voltage profile from the dataset is illustrated in Fig.~\ref{figure_experiment}. The degradation trajectory reveals complex, non-linear dynamics characterized by distinct local trends. Although environmental and operational metadata $\mathbf{m}^{(i)}$ are recorded, the metadata vector is omitted from the core modeling pipeline, reducing the objective to a purely univariate time-series forecasting task. This was motivated by a preliminary exploratory analysis, where metadata were incorporated as features and coupled to the flattened input window. These experiments showed no benefit or a decreasing of predictive performance, indicating that, in our setting, metadata static features provide only a negligible predictive signal for the temporal degradation horizon considered. 

\subsection{Preprocessing}

The raw signals are irregularly sampled, with higher acquisition frequency during transient regimes characterized by rapid signal variations, resulting in a non-uniform temporal grid. Additionally, the measurements are affected by significant high-frequency noise. To obtain a uniformly sampled and noise-robust representation suitable for learning, the following preprocessing pipeline is applied independently to each experiment:

\begin{enumerate}
    \item \textbf{Gaussian smoothing}, used to attenuate high-frequency noise while preserving long-term degradation trends.
    
    \item \textbf{Cubic interpolation:} a cubic interpolant is fitted to the smoothed signal, yielding a continuous-time function
    \begin{equation}
    \tilde V^{(i)} : [0,\mathcal T_i] \rightarrow \mathbb R.
    \end{equation}
    
    \item \textbf{Uniform resampling}, producing a discrete-time signal
    \begin{equation}
    V^{(i)}_n := \tilde V^{(i)}(n\Delta t), \quad n = 0,\dots,T_i,
    \end{equation}
    with fixed sampling interval $\Delta t = 5$ minutes.
    Note that $T_i$ may change depending on $i$, since the experiments have different duration.
\end{enumerate}

This temporal resolution reflects the fact that electrochemical degradation evolves on timescales significantly longer than one hour, making finer sampling unnecessary for the prediction task.



\subsection{Final dataset construction}

A key challenge is the strong heterogeneity across experiments, which differ in duration, voltage range, membrane and catalyst properties, and noise characteristics. To enable learning across experiments of varying length, we construct a dataset of fixed-size input-output pairs using a sliding-window approach.

\paragraph{Sliding windows}
We define the window length $L$ and stride $f$ as
\begin{equation}
L = \frac{24\ \mathrm{h}}{\Delta t}, 
\quad 
f = \frac{1\ \mathrm{h}}{\Delta t}.
\end{equation}
For each experiment, overlapping windows are extracted as
\begin{equation}
\widetilde W_{i,j} = \bigl(V^{(i)}_{n^{(i)}_j}, \dots, V^{(i)}_{n^{(i)}_j+L-1}\bigr),
\quad n^{(i)}_j = j f,
\end{equation}
where $i=1,\dots , N$ is the experiment index.

The window length corresponds to a full 24-hour cycle, allowing the models to capture daily variations in operating conditions. The chosen stride results in highly overlapping windows; to prevent data leakage, all windows extracted from a single experiment are assigned to the same split during evaluation.

\paragraph{Prediction task}

The goal is to learn a mapping
\begin{equation}
f_\theta : \mathbb R^L \rightarrow \mathbb R
\end{equation}
that predicts a scalar summary of the future system state given a past observation window.

Rather than predicting a single future value, we adopt a direct multi-horizon formulation. Let $H \in \{3,6,12,18,24\}$ denote the prediction horizon and $H_s = H/\Delta t$. The target associated with $\widetilde W_{i,j}$ is defined as
\begin{equation}
\tilde y_{i,j} = \frac{1}{A} \sum_{n=n^{(i)}_j+L+H_s-A}^{n^{(i)}_j+L+H_s-1} V^{(i)}_n,
\quad A = \frac{1}{4} H_s.
\end{equation}

This definition corresponds to averaging over the final portion of the prediction horizon, yielding a robust estimate of the future regime and mitigating the effect of high-frequency noise.

\paragraph{Normalization}

To account for distributional differences across experiments and avoid data leakage, we adopt a per-window normalization scheme. For each window $\widetilde W_{i,j}$, we compute its mean $\mu_{i,j}$ and standard deviation $\sigma_{i,j}$, and define
\begin{equation} \label{Eq:norm}
W_{i,j} = \frac{\widetilde W_{i,j} - \mu_{i,j}}{\sigma_{i,j}}, 
\quad 
y_{i,j} = \frac{\tilde y_{i,j} - \mu_{i,j}}{\sigma_{i,j}}.
\end{equation}
This local standardization emphasizes relative temporal dynamics while ensuring robustness to variations in absolute voltage scale across experiments. 

\paragraph{Final dataset}

The final dataset is
\begin{equation}
\mathcal{D}_f = \bigl\{ (W_{i,j}, y_{i,j}) \bigr\},
\end{equation}
where the experiment index $i=1,\dots,N$ is kept in order to implement the evaluation procedure we will introduce later. Note that, since each experiment $i$ has different duration, the index $j$ will range on an index set depending on $i$.

\section{Methodology and Architectures}

The forecasting task is characterized by a severe low-data regime, with the final dataset $\mathcal{D}_f$ containing approximately 200--500 samples depending on the prediction horizon. This setting introduces a high risk of overfitting and makes careful control of model complexity essential. 

To address this challenge, we adopt a strict cross-experiment evaluation protocol together with a restrained modeling strategy.

\subsection{Evaluation Protocol}

The dataset consists of $N = 12$ independent experimental runs, each generating multiple highly overlapping samples via a sliding-window procedure. Standard random splits are not appropriate in this setting, as they would mix samples from the same experiment across training and test sets, leading to data leakage and overly optimistic performance estimates.

We therefore adopt a \emph{Leave-One-Group-Out Cross-Validation} (LOGO-CV) strategy, where each experimental run defines a group. Let $\mathcal{E} = \{E_1, \dots, E_N\}$ denote the set of experiments. For each fold $k$, we define:
\begin{itemize}
    \item \textbf{Test set:} all samples derived from $E_k$;
    \item \textbf{Training set:} all samples derived from $\mathcal{E} \setminus \{E_k\}$.
\end{itemize}

Each model is trained from scratch on the training folds and evaluated on the held-out experiment. Final performance is reported as the average across all folds, providing an estimate of cross-experiment generalization, while avoiding the leakage of data between training and evaluation.

\subsection{Validation and Model Selection}

Due to the limited number of independent experiments, we do not employ an additional validation split. Preliminary experiments showed that such splits lead to high-variance performance estimates and unstable model selection, as results depend strongly on the specific choice of validation experiments.

Instead, all architectures and hyperparameters are set based on standard practice (see Section \ref{sec:models_and_architectures}). These configurations remain constant across all cross-validation folds. This restrained strategy reduces the risk of overfitting to the evaluation protocol and ensures a consistent comparison across models.

We emphasize that the objective of this study is not to achieve state-of-the-art performance through extensive hyperparameter tuning, but rather to evaluate the robustness and relative inductive biases of different modeling approaches in a low-data regime.

\subsection{Models and Architectures} \label{sec:models_and_architectures}

We consider a hierarchy of models with increasing expressive power, ranging from linear baselines to deep neural architectures. All models are implemented in Python, using \texttt{scikit-learn} for classical methods and \texttt{TensorFlow} for neural architectures.
Models are trained using the mean squared error loss; neural models are optimized with Adam \cite{b15} ($\text{lr} = 10^{-3}$), employing a learning rate scheduler to improve convergence and early stopping to prevent overfitting.

\subsubsection{Baselines}

\begin{table*}[!ht]
\caption{Global performance summary: Comparison of models across time horizons (MAE).}
\begin{center}
\begin{tabular}{|c|c|c|c|c|c|c|c|}
\hline
\textbf{Horizon} & \textbf{Metric} & \textbf{\textit{Linear}} & \textbf{\textit{Ridge}} & \textbf{\textit{RF}} & \textbf{\textit{Shallow NN}} & \textbf{\textit{LSTM}} & \textbf{\textit{1D-CNN}} \\
\hline
\multirow{2}{*}{\textbf{24h}}  
& Norm & $2.708$ & $2.567$ & $2.960 \pm 0.015$ & $\mathbf{1.816} \pm \mathbf{0.015}$ & $1.977 \pm 0.205$ & $1.845 \pm 0.158$ \\ \cline{2-8}
& cV   & $2.308$ & $1.909$ & $2.362 \pm 0.007$ & $\mathbf{1.000} \pm \mathbf{0.005}$ & $1.063 \pm 0.146$ & $1.284 \pm 0.221$ \\ \hline
\multirow{2}{*}{\textbf{18h}}  
& Norm & $2.386$ & $2.127$ & $2.251 \pm 0.023$ & $1.733 \pm 0.019$ & $\mathbf{1.730} \pm \mathbf{0.035}$ & $1.756 \pm 0.059$ \\ \cline{2-8}
& cV   & $2.001$ & $1.710$ & $1.772 \pm 0.024$ & $1.150 \pm 0.020$ & $\mathbf{1.091} \pm \mathbf{0.059}$ & $1.359 \pm 0.074$ \\ \hline
\multirow{2}{*}{\textbf{12h}}  
& Norm & $2.064$ & $1.614$ & $1.727 \pm 0.025$ & $\mathbf{1.471} \pm \mathbf{0.004}$ & $1.562 \pm 0.041$ & $1.543 \pm 0.007$ \\ \cline{2-8}
& cV   & $1.868$ & $1.393$ & $1.483 \pm 0.014$ & $1.180 \pm 0.010$ & $\mathbf{1.157} \pm \mathbf{0.039}$ & $1.278 \pm 0.013$ \\ \hline
\multirow{2}{*}{\textbf{6h}}   
& Norm & $1.612$ & $1.020$ & $1.042 \pm 0.011$ & $\mathbf{0.970} \pm \mathbf{0.009}$ & $1.154 \pm 0.035$ & $1.025 \pm 0.009$ \\ \cline{2-8}
& cV   & $1.566$ & $0.867$ & $0.908 \pm 0.009$ & $\mathbf{0.850} \pm \mathbf{0.010}$ & $0.965 \pm 0.028$ & $0.887 \pm 0.010$ \\ \hline
\multirow{2}{*}{\textbf{3h}}   
& Norm & $1.251$ & $\mathbf{0.644}$ & $0.683 \pm 0.002$ & $0.675 \pm 0.003$ & $0.790 \pm 0.045$ & $0.668 \pm 0.002$ \\ \cline{2-8}
& cV   & $1.235$ & $\mathbf{0.575}$ & $0.607 \pm 0.002$ & $0.620 \pm 0.000$ & $0.690 \pm 0.036$ & $0.587 \pm 0.010$ \\ \hline
\end{tabular}
\label{tab:mae_combined_multirow}
\end{center}
\end{table*}

\paragraph{Global linear trend}
As a simple heuristic, we assume that the degradation trend observed in the input window remains constant over the prediction horizon. For each sample, we fit
\begin{equation}
v(t) = \alpha t + \beta,
\end{equation}
and extrapolate to the target time $t_{\text{target}}$:
\begin{equation}
\hat{y} = \alpha t_{\text{target}} + \beta.
\end{equation}

\paragraph{Ridge regression}
We model the target as a linear function of past observations:
\begin{equation}
\hat{y} = \mathbf{w}^\top \mathbf{x} + b,
\end{equation}
where $\mathbf{x}$ is the flattened input window. Parameters are estimated using ridge regression with $L^2$ regularization ($\alpha = 1$), ensuring stability under strong collinearity in the input features.

\paragraph{Random Forest}
We include a Random Forest regressor as a strong non-parametric baseline for tabular representations of the input windows. The model is configured with 300 trees, minimum leaf size of $5$, and $\sqrt{L}$ feature subsampling, where $L=288$ is the input dimensionality. This configuration provides a balance between variance reduction and robustness in low-sample regimes.

\subsubsection{Shallow neural network}

We consider a shallow fully connected ReLU Neural Network, implemented with strong regularization. The architecture consists of a single hidden layer with 128 ReLU units, $L^2$ normalization, and dropout (rate 0.2).

\subsubsection{Deep temporal models}

\paragraph{1D Convolutional Neural Network (1D-CNN)}
The 1D-CNN captures local temporal dependencies through convolutional filters. The architecture consists of a single convolutional layer with 16 filters and kernel size 5, followed by max pooling (pool size 2), dropout (rate 0.3), and a fully connected layer with 16 ReLU units. This design reduces parameter count while preserving sensitivity to local structure in the time series.

\paragraph{Long Short-Term Memory network (LSTM)}
The LSTM explicitly models sequential dependencies through recurrent dynamics. The input sequence is processed and summarized via the final hidden state. The model consists of a single LSTM layer with 8 units and dropout (rate 0.2), followed by a fully connected layer with 4 ReLU units and dropout (rate 0.2), and a final linear output layer.

Although more expressive in principle, the LSTM is intentionally kept small due to the low-data regime. Empirically, increasing model capacity led to degraded performance, suggesting strong overfitting sensitivity.

\section{Experimental Results and Discussion}

\subsection{Performance metrics}

Model performance is evaluated using the Mean Absolute Error (MAE). Given ground truth targets $y_i$ and predictions $\hat{y}_i$, the MAE is defined as
\begin{equation}
\text{MAE} = \frac{1}{M} \sum_{i=1}^M |y_i - \hat{y}_i|.
\end{equation}

We report MAE both in normalized units and in physical units (centiVolts, cV), obtained by inverting the per-window normalization (see Eq. \ref{Eq:norm}), along with their standard deviations across five different training and evaluation runs. Reporting results in physical units facilitates interpretation in the context of electrochemical performance and degradation.

\subsection{Results}

We present results across five different time horizons (3, 6, 12, 18, 24 hours) in Table \ref{tab:mae_combined_multirow}. The best performing baseline is Ridge Regression by a clear margin: the global linear trend struggles to capture the full complexity of the degradation phenomenon, while the Random Forest shows weaker performance at longer horizons. 

The Shallow Neural Network achieves the most consistent performance across all time scales, as well as having lower standard deviation between runs, compared to temporal models, which highlights more robust predictions. Regarding deep temporal models, the LSTM performs well on longer time horizons but shows degraded performance on shorter time windows, while the 1D-CNN provides stable results across all horizons, with slightly higher error than the Shallow Neural Network.

Regarding the centiVolts MAE, the errors remain consistently bounded across all forecasting horizons, without exhibiting a systematic increase at longer time scales. This suggests that the models are learning physically meaningful signal variations rather than overfitting short-term noise, and that the improvement in normalized performance translates into stable accuracy in the original voltage scale.

\section{Conclusions}

The proposed framework proves robust in handling heterogeneous and data-scarce conditions, supporting its applicability in industrial settings. In this regime, simple and strongly regularized models such as the Shallow Neural Network achieve the most consistent performance across time horizons. The 1D-CNN provides competitive results, indicating that convolutional temporal biases are effective across different forecasting ranges. In contrast, the LSTM shows less stable performance, particularly on shorter horizons, and appears more sensitive to capacity and training choices.

Overall, the results suggest that, under severe data scarcity and noisy electrochemical time series, simple regularized neural architectures offer the most reliable trade-off between performance and stability in our setting.

\section*{Acknowledgements}
Funded by the European Union. This work is supported by the Horizon Europe Grant Agreements No.~101251004 (NEREUS) and No.~101251223 (SCALE-AEM), through the Clean Hydrogen Partnership and its members. Views and opinions expressed are however those of the author(s) only and do not necessarily reflect those of the European Union or the Clean Hydrogen Joint Undertaking. Neither the European Union nor the granting authority can be held responsible for them. This work was also supported by the Italian Ministry of the Environment and Energy Security (MASE) through the Mission Innovation 2.0 project ``Sole di notte'' (ID: MI\_ERE\_00192), CUP: F33C25001220001.

\vspace{12pt}

\end{document}